\documentclass[letterpaper]{article} % DO NOT CHANGE THIS
\usepackage{aaai25_with_hyperref}  % DO NOT CHANGE THIS
\usepackage{times}  % DO NOT CHANGE THIS
\usepackage{helvet}  % DO NOT CHANGE THIS
\usepackage{courier}  % DO NOT CHANGE THIS
\usepackage[hyphens]{url}  % DO NOT CHANGE THIS
\usepackage{graphicx} % DO NOT CHANGE THIS
\usepackage{natbib}  % DO NOT CHANGE THIS AND DO NOT ADD ANY OPTIONS TO IT
\usepackage{caption} % DO NOT CHANGE THIS AND DO NOT ADD ANY OPTIONS TO IT
\usepackage{algorithm}
\usepackage{algorithmic}
\usepackage{xcolor}
\usepackage{hyperref}

\usepackage{newfloat}
\usepackage{listings}
\DeclareCaptionStyle{ruled}{labelfont=normalfont,labelsep=colon,strut=off} % DO NOT CHANGE THIS
\floatstyle{ruled}
\newfloat{listing}{tb}{lst}{}
\floatname{listing}{Listing}
\usepackage{url}
\usepackage{ifthen}
\usepackage{subfigure}
\usepackage{xcolor}
\usepackage{todonotes}
\usepackage{graphicx}

\newcommand{\UT}{The University of Texas at Austin}
\newcommand{\UTshort}{UT Austin}
\definecolor{student}{RGB}{31,119,180}
\definecolor{auditor}{RGB}{255,126,16}

\title{The Essentials of AI for Life and Society: \\An AI Literacy Course for the University Community}
\author{
  Joydeep Biswas\thanks{The author order includes the three co-instructors of the course in alphabetical order by last name, followed by the other co-authors, also in alphabetical order.
  },
  Don Fussell,
  Peter Stone,\\
  Kristin Patterson,
  Kristen Procko,
  Lea Sabatini,
  Zifan Xu
}
\affiliations{
    The University of Texas at Austin\\
    2317 Speedway, Stop D9500, Austin, TX 78712
}

\begin{document}
\maketitle

\begin{abstract}
We describe the development of a one-credit course to promote AI literacy at \UT{}. In response to a call for the rapid deployment of class to serve a broad audience in Fall of 2023, we designed a 14-week seminar-style course that incorporated an interdisciplinary group of speakers who lectured on topics ranging from the fundamentals of AI to societal concerns including disinformation and employment. University students, faculty, and staff, and even community members outside of the University, were invited to enroll in this online offering: \textit{The Essentials of AI for Life and Society}. We collected feedback from course participants through weekly reflections and a final survey. Satisfyingly, we found that attendees reported gains in their AI literacy. We sought critical feedback through quantitative and qualitative analysis, which uncovered challenges in designing a course for this general audience. We utilized the course feedback to design a three-credit version of the course that is being offered in Fall of 2024. The lessons we learned and our plans for this new iteration may serve as a guide to instructors designing AI courses for a broad audience.
\end{abstract}

\section{Introduction}

%Course objectives - the need for the course
The increased public availability of large language model-based chat tools prompted a burst of interest in artificial intelligence (AI) in the higher education community in 2023. Despite the wealth of information about AI available in the literature and through blog posts, most of it was geared toward a technical audience. News stories tended to focus on recent innovations with sweeping claims about how AI would change life and society, without providing background on how AI tools work and their limitations.

Responding to the interest in and concern about AI by faculty, staff, and students across disciplines at \UT{}
%(\UTshort{})
, the Dean of the College of Natural Sciences requested the rapid development of a new AI course targeting a non-technical audience. With a mere six weeks remaining until the start of the fall semester, we assembled a team that included three computer science faculty leads, a project manager, and experts in instructional design and assessment to create a broadly accessible new course about AI. The institutional support and multidisciplinary team were critical to the swift course deployment.

In the years leading up to the development of this course, \UTshort{} had already built
institutional structures that allowed rapid responses to emerging research and
education needs. \UTshort{} recognized the need to understand and shape the sociotechnical
challenges innate to developing and deploying AI-enabled systems in the real world. In 2019,
\UTshort{} launched Good Systems\footnote{\url{https://bridgingbarriers.utexas.edu/good-systems}}, a research grand challenge to form human-AI partnerships that are
beneficial to society. Good Systems comprises an interdisciplinary team of faculty,  which
promotes research in ethical AI use by leveraging its benefits while considering and mitigating
harms or unintended consequences that may be caused by new technologies. Conveniently, Good Systems was well established when the course was proposed, and its members supported the computer
science faculty in responding to the call.
Another key source of institutional support was a college-level STEM education center, whose
flexible structure, pedagogical and course-design expertise, and experience navigating complex
institutional systems enabled rapid development and deployment of the course.

% Another source of institutional support came from a college-level STEM education center that has a flexible structure for support of rapid change efforts. The STEM education center brought expertise in pedagogy and course design, and experience navigating complex institutional structures to promote rapid, creative change at scale.

% A third form of institutional support the proved essential for project success is a center focused on academic technology. The academic technology center had structures in place for creating online courses at scale, such as start to finish video services and a professional recording studio.

This class was designed to provide a gentle introduction to Artificial Intelligence: what it is, its
capabilities, its limitations, and its potential implications for society. To promote accessibility,
we proposed a 1-credit-hour course that students and faculty alike could easily fit into their
schedules. The course was publicized through email, department newsletters, the campus newspaper,
and social media.
Undergraduate students enrolled in the
course for credit, while graduate students, post-doctoral fellows, faculty, and staff participated as auditors.

The set of fourteen lectures on fundamental concepts for AI literacy, as well as the ethical and societal implications of AI technologies, were given by faculty from across the University. Recordings of these lectures are now available for anyone to enjoy and learn from at any time.\footnote{\url{https://www.cs.utexas.edu/~pstone/Courses/109fall23/}} Only a curiosity and willingness to explore are needed - no prior technical knowledge is required.

% \url{https://www.cs.utexas.edu/~pstone/Courses/109fall23/}

\section{Related Work}
\paragraph{AI Literacy}
There have been a few attempts to create \emph{AI Literacy} courses at universities. For example,
Kong et al.~\cite{kong2021evaluation} designed a seven-hour AI literacy course for university students from
diverse backgrounds in Hong Kong.

\paragraph{AI in K-12 Education}
While there is broad interest in inclusion of AI at the K-12 level, there is a lack of consensus on
what AI literacy means~\cite{ng2021conceptualizing}, particularly for K-12 students. Early
attempts by Ng et al.~\cite{ng2021ai} define ``AI Literacy'' as consisting of four components:
1) know and understand, 2) use, 3) evaluate, and 4) ethical issues.
To provide guidance on how
to create AI literacy curricula for K-12 students, Ng et al.~\cite{ng2023review} conducted
a
systematic review of the literature on AI literacy, producing pedagogical models, teaching tools,
and challenges as a result. Williams~\cite{williams2023review} also conducted a review of AI
literacy in K-12 curricula, with a focus on assesments, yielding a set of recommendations for
approaches to use, and existing gaps in assesment.

\paragraph{AI in University Curricula}
AI for computer science or engineering majors is a common offering at universities at the
undergraduate and graduate levels. There have also been several recent AI majors introduced recently
at the undergraduate~\cite{cmu-ai} and graduate level~\cite{utexas_ms_ai}.
There have also been several instances of integrating ethical considerations in
AI courses~\cite{vekhter2023responsible,ai_ethics_course_survey}.
\emph{CPSC 170 – AI For Future Presidents}~\cite{yale-ai} at Yale University is the closest previous
analogue
to our course, designed for all students, with no pre-requisites, and requiring no math or
programming background.

\paragraph{Relation to Previous Work}
In this paper, we present the design and retrospective analysis from offering an AI literacy course to all members of a large public university, without assumptions of technical background. Unlike previous courses, this offering is simultaneously \emph{technical} --- it covers the main technical sub-disciplines that comprise AI; while including \emph{ethical} considerations; and is aimed at a \emph{broad audience, including non-technical majors}.

% Davy Tsz Kit Ng is a Hong-Kong-based visiting scholar who keeps coming up in my searches, and seems to be focused on AI and tech pedagogy. This first article is a well-researched literature review I think we probably should cite. Then, there are a couple on defining AI literacy I’m including for your consideration.

% This is a recent review article that surveys the evolution of AI courses, and summaries the pedagogical tools used in teaching about AI.

% Ng, D. T. K., Lee, M., Tan, R. J. Y., Hu, X., Downie, J. S., & Chu, S. K. W. (2023). A review of AI teaching and learning from 2000 to 2020. Education and Information Technologies, 28(7), 8445-8501.

% These two are possibilities. Both look to the literature to give working definitions of AI literacy based on published learning goals. The first paper is cited more than the second..

% Ng, D. T. K., Leung, J. K. L., Chu, S. K. W., & Qiao, M. S. (2021). Conceptualizing AI literacy: An exploratory review. Computers and Education: Artificial Intelligence, 2, 100041.
% Ng, D. T. K., Leung, J. K. L., Chu, K. W. S., & Qiao, M. S. (2021). AI literacy: Definition, teaching, evaluation and ethical issues. Proceedings of the Association for Information Science and Technology, 58(1), 504-509.

% I also wondered if this might be a good reference to use because the first iteration did have the goal of being broadly and publicly available, but I can only access the abstract:  https://dl.acm.org/doi/abs/10.1145/3545947.3573292

\section{Course Design}
% \jb{This section could do with some structure.}
% \commentp{I could suggest the following.  "In summary, the students were expected to completed the following each week:
% \begin{itemize}
%     \item Complete assigned readings prior to the lecture;
%     \item Attend lecture;
%     \item Submit written reflection on the reading and lecture.
% \end{itemize}
% " \\
% But I"m not sure how much that adds.  Maybe at the end of the section?
% }
The ambition of the course was to introduce AI to \emph{anyone} with interest. To reach this large
target audience, accessibility was crucial. Accordingly, this online course was delivered with
support from the Liberal Arts Instructional Technology Services (LAITS) at \UTshort{}.

\subsection{Curricular Objectives}
The main learning objective of the course was to improve participants' AI literacy. To support
students' achievement of this AI literacy objective, we introduced the types of AI and their
applications, highlighted the risks and benefits of various AI technologies, discussed the societal
impacts of AI technologies, and aimed to improve learners' ability to distinguish AI science from
science fiction. The course faculty determined that background on concepts that are foundational to
AI would be included in the first five lectures, and the foundational concepts would be connected to
specific applications (computer vision, robotics, and natural language processing, including large
language models) in the following three presentations (Table~\ref{tab:schedule}). The remaining lectures provided
historical context, followed by an exploration of some of AI's inherent dangers and limitations.
Lecturers were primed to expect a wide range of backgrounds and experiences among the students, and
were asked to assume no prior technical knowledge --- as a consequence, where necessary, lecturers
included introductory technical primers in the lecture content to support understanding.

\subsection{Course Structure}
Students were expected to attend weekly one-hour lectures synchronously, and recordings were made available afterwards to provide flexibility for faculty and staff whose real-time attendance may be challenging. A limited number of students could attend the lectures in-person in the studio.

The course requirements for students were to complete readings, attend the online lectures, and
respond to in-class quiz questions on the course material. Weekly written reflections on the reading
and lecture were also required; because we were developing this course concurrently with its
delivery, we used these reflections as an opportunity to evaluate the course and make corrections.
Grades were based on attendance (30 percent), correctness of responses on real-time quizzes (40
percent), and completion of weekly reflections (30 percent). Assignments were designed to be an
incentive to participate, rather than to be evaluative --- we determined early on that it would be
acceptable, indeed desirable, for every student to earn an A.

Each lecture included about 30-40 minutes of presentation time, where one of the course faculty or a
guest lecturer presented on a designated topic. We recruited experts from computer science to
introduce the foundational concepts we had thoughtfully curated, and in the latter half of the
course we identified guest speakers from diverse departments across \UTshort{}. The guest lecturers
were instructed to tie their lectures back to a working definition of AI, but were free to structure
the talk as they liked. Each speaker selected a required reading assignment to accompany their
lecture, and was welcome to provide additional optional readings for students who wished to explore
a topic more deeply.

\begin{table*}[htb]
    \centering
    \resizebox{16cm}{!}{
    \begin{tabular}{|c|c|} \hline
         \textbf{Topic}&  \textbf{Lecturer (Department)}\\ \hline
         Introduction - AI100 study\cite{ai100} &  Peter Stone (Computer Science)\\ \hline
         Planning and Search&  Joydeep Biswas (Computer Science)\\ \hline
         Probabilistic Modeling &  Roberto Martín-Martín (Computer Science)\\ \hline
         Machine Learning Fundamentals &  Adam Klivans (Computer Science)\\ \hline
         Machine Learning Paradigms &  Ray Mooney (Computer Science)\\ \hline
         Computer Vision &  Kristen Grauman (Computer Science)\\ \hline
         Intelligent Robotics &  Luis Sentis (Aerospace Engineering /Engineering Mechanics)\\ \hline
         Natural Language Processing (Large Language Models) &  Greg Durrett (Computer Science)\\ \hline
         Philosophical Foundations and Relation to Computing &  Don Fussell (Computer Science)\\ \hline
 AI and Mis/disinformation & Matt Lease (School of Information)\\ \hline
 Bias and Fairness in AI Models / Elargethical Datasets & S. Craig Watkins (Journalism and Media)\\ \hline
 Impacts on Workplace and Economics& Sherri Greenberg (LBJ School of Public
Affairs)\\\hline
 AI Alignment and Existential Threats & Scott Aaronson (Computer Science)\\\hline
 Current and Future Directions  & Joydeep Biswas, Don Fussell, Peter Stone (Computer Science)\\\hline
    \end{tabular}
    }
    \caption{Course lecture schedule}
    \label{tab:schedule}
    \vspace{-10pt}
\end{table*}

% \begin{table*}[htb]
%     \centering
%     \begin{tabular}{|c|c|} \hline
%          \textbf{Topic}&  \textbf{Lecturer (Department)}\\ \hline
%          Introduction - AI100 study\cite{ai100} &  Paper Author 1 (Computer Science)\\ \hline
%          Planning and Search&  Paper Author 2 (Computer Science)\\ \hline
%          Probabilistic Modeling &  Redacted (Computer Science)\\ \hline
%          Machine Learning Fundamentals &  Redacted (Computer Science)\\ \hline
%          Machine Learning Paradigms &  Redacted (Computer Science)\\ \hline
%          Computer Vision &  Redacted (Computer Science)\\ \hline
%          Intelligent Robotics &  Redacted (Aerospace Engineering /Engineering Mechanics)\\ \hline
%          Natural Language Processing (Large Language Models) &  Redacted (Computer Science)\\ \hline
%          Philosophical Foundations and Relation to Computing &  Paper Author 3 (Computer Science)\\ \hline
%  AI and Mis/disinformation & Redacted (School of Information)\\ \hline
%  Bias and Fairness in AI Models / Ethical Datasets & Redacted (Journalism and Media)\\
%  Impacts on Workplace and Economics& Redacted (LBJ School of Public
% Affairs)\\\hline
%  AI Alignment and Existential Threats & Redacted (Computer Science)\\\hline
%  Current and Future Directions  & Paper Authors 1,2,3 (Computer Science)\\\hline
%     \end{tabular}
%     \caption{Course lecture schedule}
%     \label{tab:schedule}
% \end{table*}

\subsection{Evaluations and Learning Assesments}
Students were asked to complete the associated reading assignments prior to watching the lectures. To promote engagement in the synchronous online lecture, we incorporated live quiz questions. We originally interspersed the questions throughout the presentation via a polling system to encourage attentiveness in the largely remote audience; however, midway through the course we opted to change the format to provide a single integrated quiz following the lecture. This mitigated some technical difficulties and gave the speaker an opportunity to transition to a seat beside the host faculty member to engage in 5-10 minutes of Q\&A about the presentation. Course participants submitted questions through a chat feature, with priority given to questions from students in the live studio audience.

% Reading assignments can be found here:  \url{https://www.cs.utexas.edu/~pstone/Courses/109fall23/assignments.html}

% Lecture recordings can be found here:  \url{https://www.youtube.com/playlist?list=PLJYKxIV9Xh7RRam6LFJJb5uFJ7-19u7j4}

\section{Student Reactions}

We analyze the effectiveness of the course in terms of 1) the background of the enrolled students, 2) responses to weekly surveys, and 3) a final course survey.

\subsection{Enrollment}
The course was successful in enrolling a broad audience. A total of 788 individuals signed up for the course, including 132 undergraduates who signed up for credit, 631 auditors from the university, and 25 external participants.
Among the 631 university individuals that signed up to audit, all 17 colleges from \UTshort{} were represented. University-affiliated auditors included faculty, emeritus faculty, staff, postdoctoral fellows, graduate students, and alumni. Table~\ref{tab:enrollment} lists the college affiliations and standings of the university individuals who enrolled for the course.
% \kp{Do we want to expand at all here? Also, we need to figure out exactly how we want to report this: I added a new document to survey folder on Box. A total of 656 auditors signed up. In the second tab, I totaled UT folks by college, but numbers were short. Below the table is corrected numbers with free response items folks typed in and blanks, which indicated the person was external (they filled in their affiliations in another column on the sheet). On Canvas, 584 individuals actually logged into the course. This is the number I included in the report for our February meeting, but I don't have a good way to disaggregate them from the 656 total signed up. I think this is the cleanest way to describe it but open to suggestions}
\begin{table}[htb]
\centering
\resizebox{7.5cm}{!}{
\begin{tabular}{|l|r|r|r|r|r|r|}
\hline
\textbf{UNIT} & \textbf{UG} & \textbf{GS} & \textbf{P} & \textbf{F} & \textbf{S} & \textbf{T} \\
\hline
Engineering & 3 & 22 & 3 & 9 & 6 & 40 \\
Education & 1 & 3 & 0 & 28 & 33 & 64 \\
Fine Arts & 2 & 1 & 0 & 10 & 7 & 18 \\
Liberal Arts & 36 & 12 & 1 & 43 & 11 & 67 \\
Natural Sciences & 80 & 24 & 3 & 67 & 51 & 145 \\
Pharmacy & 0 & 1 & 1 & 1 & 0 & 3 \\
Medical School & 0 & 2 & 1 & 2 & 25 & 30 \\
Geosciences & 0 & 0 & 0 & 0 & 1 & 1 \\
Public Affairs & 0 & 0 & 0 & 2 & 0 & 2 \\
Business & 9 & 1 & 0 & 10 & 13 & 24 \\
Communication & 1 & 4 & 1 & 16 & 2 & 23 \\
Architecture & 0 & 1 & 0 & 9 & 6 & 16 \\
Information & 0 & 22 & 3 & 0 & 1 & 26 \\
Law & 0 & 1 & 0 & 4 & 1 & 6 \\
Nursing & 0 & 2 & 0 & 0 & 1 & 3 \\
Social Work & 0 & 0 & 0 & 3 & 7 & 10 \\
UG College & 0 & 0 & 0 & 0 & 8 & 8 \\
Other & 0 & 5 & 0 & 12 & 129 & 146 \\
\hline
% \textbf{Total of Type} & 132 & 96 & 13 & 204 & 173 & 787 \\
\textbf{Total of Type} & 132 & 101 & 13 & 216 & 302 & 763 \\
\hline
\end{tabular}
}
\caption{Course enrollment by unit and standing: undergraduate students (UG), graduate students (G), postdocs (P), faculty (F), staff (S), and unit total (T).}
\label{tab:enrollment}
\vspace{-10pt}
\end{table}
\subsection{Response to Weekly Surveys}

% We evaluated the course in two ways: 1. through targeted questions incorporated with the weekly reflections, and 2. through a final course survey.

% Survey folder: \url{https://utexas.app.box.com/folder/277865200543}

The weekly reflections were a required assignment for students that was graded based on completion; auditors were encouraged, but not required, to complete weekly reflections. Each week, we collected over 100 student responses and usually 10–20 reflections from course auditors.
% \kp{The first two weeks we had more responses from auditors but I am reluctant to highlight the drop off}
Following a brief free-form essay reflecting on the course material, we integrated several rating
scales to gauge the amount of time students required to complete the reading, their level of
engagement with the topic, and prior understanding. While designing the course, we anticipated
offering a 3-credit option in the future. Accordingly, we included one open-ended targeted question:
``If this topic were expanded, what would you like to learn more about? In the space below, list any
relevant subtopics you would like more information about.'' In addition to using these responses to
identify future subtopics to expand upon, we uncovered challenges students faced with some readings
and course material. In general, student responses indicated a desire for more examples connecting
the early lectures on fundamentals to specific applications in AI, which was largely limited by the
single, brief lecture period. They also requested current news articles to connect topics to the
most recent advances in the field. Some weeks, responses to the open-ended question revealed the
reading selected by the guest speaker or the lecture content presented was too high-level for the
broad audience.

\begin{figure*}[t]
    \centering
    \subfigure[How much of the course did you engage in?]{\includegraphics[width=0.3\linewidth]{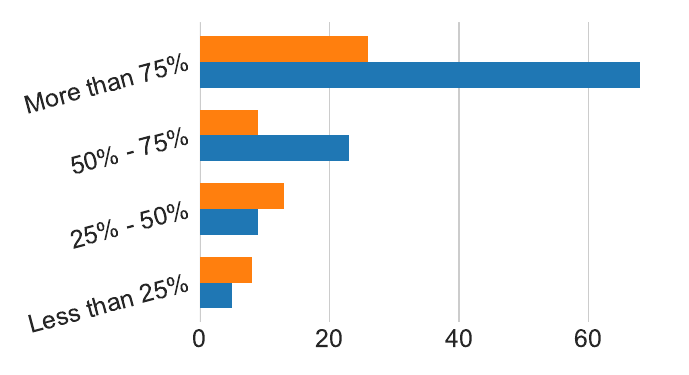}}
    \subfigure[Please rate this course on how interesting it was:]{\includegraphics[width=0.3\linewidth]{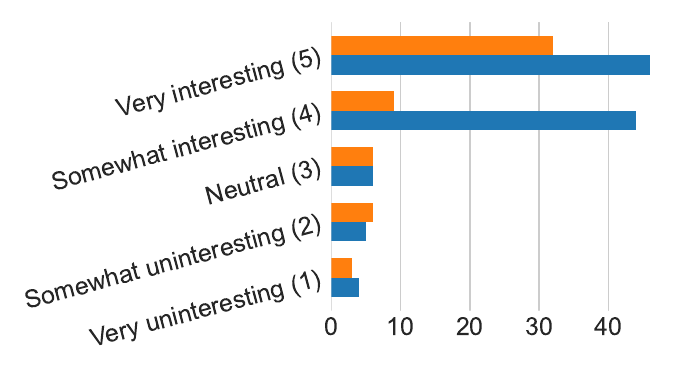}}
    \subfigure[Please rate this course on how useful it was to you:]{\includegraphics[width=0.3\linewidth]{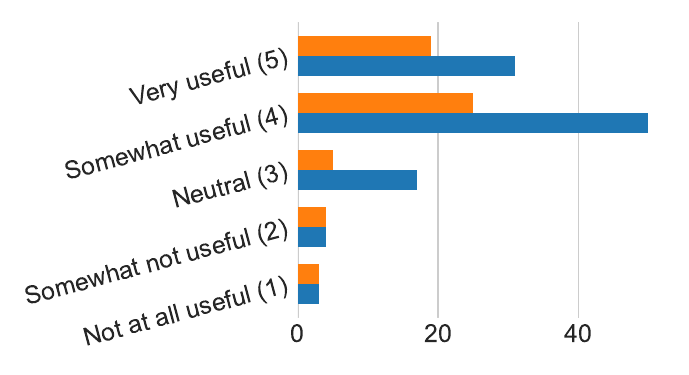}}

    \subfigure[How easy/difficult did you find the readings overall?]{\includegraphics[width=0.3\linewidth]{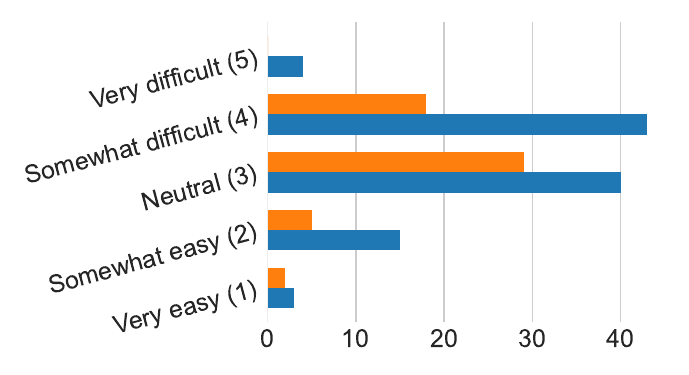}}
    \subfigure[How easy/difficult did you find the lecture material overall?]{\includegraphics[width=0.3\linewidth]{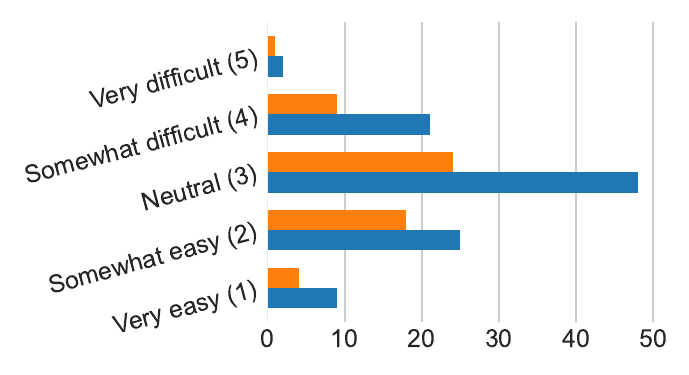}}
    \subfigure[How likely or unlikely are you to recommend this course to another person?]{\includegraphics[width=0.3\linewidth]{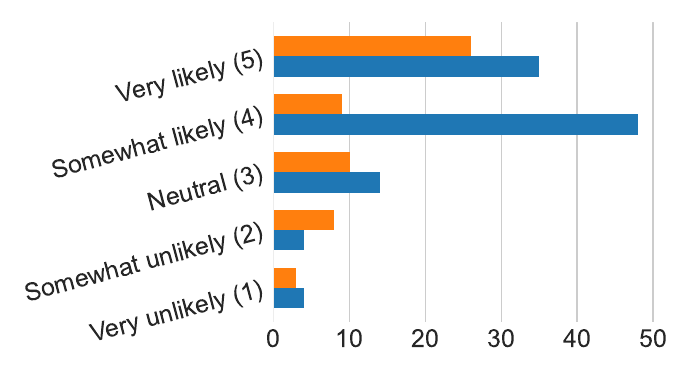}}

    \subfigure[This course cleared up a misconception I had about AI.]{\includegraphics[width=0.3\linewidth]{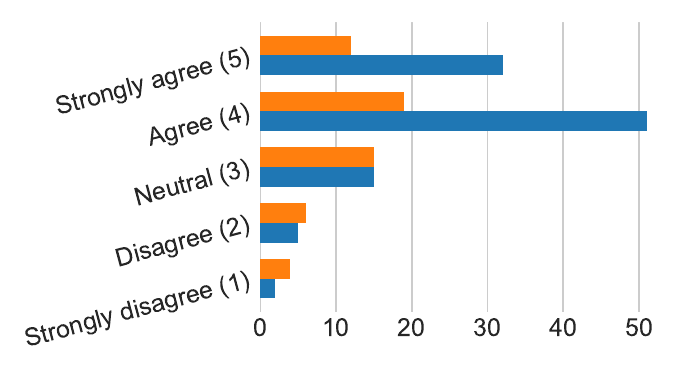}}
    \subfigure[This course helped me understand how AI may impact me in my specific field.]{\includegraphics[width=0.3\linewidth]{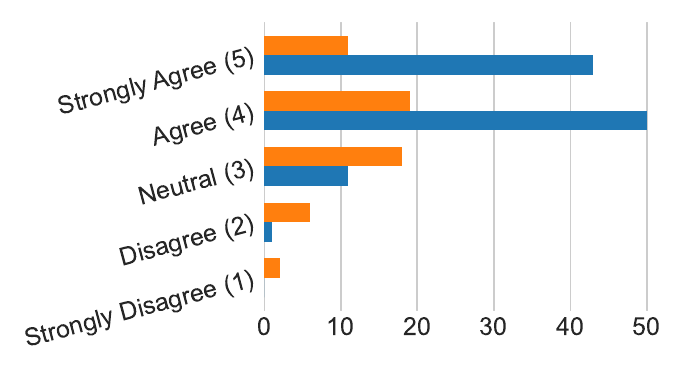}}
    \subfigure[I learned more from this course than I could have on my own if I had actively sought out the material.]{\includegraphics[width=0.3\linewidth]{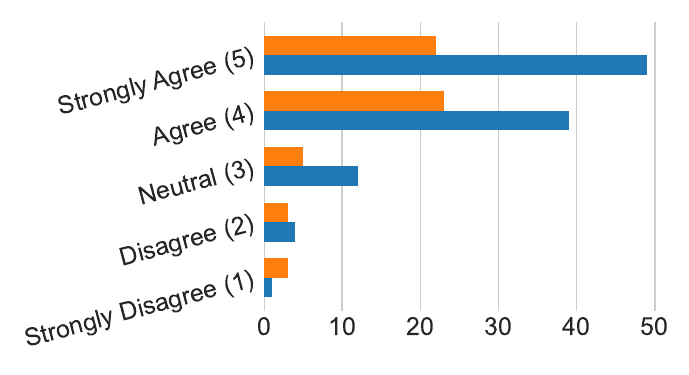}}

    \caption{Summary of survey responses on engagement and interest, difficulty of course materials, and overall understanding. The \textcolor{student}{blue} and \textcolor{auditor}{orange} bars represent responses from students and auditors, respectively. Subtitles indicate the prompts corresponding to each survey question.}
    \label{fig:course_engagement}
    \vspace{-10pt}
\end{figure*}

\subsection{Overall Course Survey}

The UT Course Evaluation Survey (CES) includes a 1-5 Likert scale for the overall course, and the
course instructor(s). Our course ratings were 4.23 and 4.33, respectively. For the same semester,
the average ratings for all courses in the College of Natural Sciences were 4.00 and 4.20. In addition to the standard CES, we conducted an extensive evaluation to assess the course's effectiveness
at meeting its curricular objectives, which we present next. Unfortunately, unlike the CES, there are no baselines to compare the detailed evaluation results to, but we hope
our results will serve as baselines for evaluations of future iterations at UT and/or elsewhere.

We replaced the final reflection essay with a targeted course survey which was part of the students' course grade. To recruit auditors to complete the survey, we sent several targeted messages to this group through the learning management system. We encouraged them to respond to the survey regardless of the amount of the course they had completed. We collected responses from 145 course participants, the majority (70 percent) of whom were undergraduate students. The remaining respondents were auditors: 12 percent were faculty, 11 percent were staff, and the remaining 7 percent were graduate students or indicated other roles at \UTshort{}. Among those internal to \UTshort{}, the majority (54 percent) were affiliated with the College of Natural Sciences, with an additional 24 percent from the College of Liberal Arts and 8 percent from the McCombs
School of Business.
Figure~\ref{fig:course_engagement} summarizes the survey responses.

We first evaluated the data for course engagement. Gratifyingly, the majority of respondents (73 percent) indicated that they were likely to recommend the course to another person. Regarding how interesting participants found the course, 81 percent of respondents indicated that the course was "very interesting" or "somewhat interesting" (average rating of 4.14 on a 5-point Likert scale where 1 = very uninteresting and 5 = very interesting).
% \kp{This includes responses of somewhat likely and very likely}
Perhaps unsurprisingly, students who were earning a grade in the course were more engaged than auditors: 87 percent of students reported that they engaged with more than half of the material, compared to 63 percent of auditors. We also evaluated engagement through attendance logs, which peaked in the first week at 387 live participants and in later weeks dropped to 170.
% \kp{not including the final week of the course, which was optional for many students that had met attendance requirements. That number was 152.}

Open-ended questions were also used to evaluate the course. Response to the question: ``What
aspect(s) of the course did you find most useful?" were thematically
coded~\cite{auerbach2003qualitative,saldana2021coding} to uncover themes in participant responses.
Of the 122 open-ended responses collected, 22 percent described the variety of guest speakers and
broad overview.  
%, for example: ``I loved that we heard from different guest speakers every week; it
%was great to get to see so many experts in so many different fields. I also found the course content
%to be sufficiently reflective of the instructors' thesis that the AI problem is an interdisciplinary
%problem, not just a computer science problem as is commonly thought of." 
The lectures stood out as a
useful learning tool with 37 percent of open-ended responses mentioning them. We also sought
critical feedback through the open-ended question:  ``What aspect(s) of the course needs
improvement?'' Most of the content-related comments (19 percent of 115 open-ended responses) focused
on the level and length of the reading, for example: ``In my opinion, some of the reading was
mathematically challenging, so I would recommend adding prerequisite courses for students." This was
corroborated by quantitative analysis, where survey respondents were asked to rate the question:
``How easy/difficult did you find the readings overall?" on a five-point Likert scale (1=very easy
to 5=very difficult). Of the 159 responses, a majority (51 percent) selected ``somewhat difficult"
or ``very difficult". In an analogous question about the reading, only 30 percent rated the lecture
material as difficult. Other course structure critiques emerging in the open-ended feedback (13
percent) focused on interactivity and specific assignments: ``I would like to see an active student
community, either making the lectures way more interactive, or implementing new and fun ideas for
students to participate in.''

\begin{figure}
    \centering
    \includegraphics[width=0.9\linewidth, trim=10 10 10 10, clip]{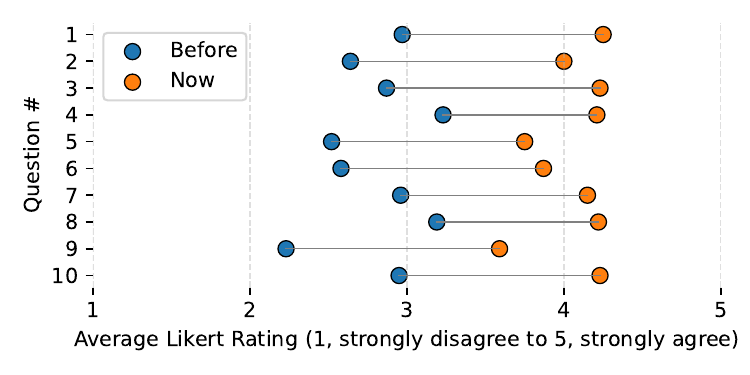}
    \caption{Retrospective pre-/post-survey questions. Each result is statistically significant, with a p-value $<$ 0.01, n = 151 for all questions, except for Q5 where n = 150.}
    \label{fig:survey}
\vspace{-10pt}
\end{figure}

Our overarching goal for this course was to improve AI literacy for a broad audience. We utilized retrospective pre-post survey~\cite{geldhof2018revisiting,howard1979response} questions to examine this; through the course survey given at the end of the course, participants asked to rate their skill level before the course and now, after taking the course on a 5-point Likert scale (1 = strongly disagree and 5 = strongly agree). For each of ten questions probing participants' AI literacy, we saw an increase in the rating after taking the course. The list of questions follows; in parenthesis following the question the change in rating is reported. Each reported change is statistically significant with a \textit{P}-value $< 0.01$ and $n = 151$ for all questions (except Q5 where $n = 150$). Figure~\ref{fig:survey} summarizes the average Likert rating for each question in the survey.
\begin{enumerate}
{\small
    \item I can define artificial intelligence (AI). (+1.27)
    \item I have the necessary vocabulary to discuss AI. (+1.36)
    \item I can list five examples of AI. (+1.37)
    \item I can describe how AI affects my daily life. (+0.97)
    \item My understanding of AI makes me well-equipped to apply it to my future professional work. (+1.24)
    \item I am prepared to weigh in on the deployment of AI in products that affect me. (+1.29)
    \item I can evaluate news stories about AI. (+1.19)
    \item I can differentiate AI science from science fiction. (+1.03)
    \item I am literate about the technical components of AI. (+1.36)
    \item I am literate about the societal implications of AI. (+1.28)
}
\end{enumerate}
% \subsection{Evaluating Effectiveness Of The Course}

\section{Lessons Learned}

Analysis of course feedback shed light on key considerations when designing an AI literacy course for a broad audience. Course participants appreciated the variety of speakers and connection to examples. Indeed, participants expressed a seemingly limitless desire for additional examples in the weekly reflection feedback. Interest in the integration of popular news articles connected to the weekly topics was noted frequently in the open-ended reflections as well. These types of articles could mitigate the main course flaw, which was that speaker-selected readings were often challenging for our non-technical audience.

Another key takeaway is that a broad audience from different backgrounds can be difficult to satisfy. Auditors reported less course engagement and some described having different expectations for the course (e.g., wanting to learn how to use AI tools, a greater focus on incorporating AI in teaching). Some staff members with limited experience using the University learning management system reported difficulty navigating and finding course materials. Gratifyingly, despite these challenges, the audience that participated in the final course survey improved their AI literacy.

\subsection{Summary Of Lessons Learned}

\paragraph{Engaging students and a wide variety of speakers can pose challenges for an AI literacy course.}
Enrollment in the course included 131 students and 584 auditors, which included faculty, staff, postdoctoral fellows, graduate students, and participants external to \UT{}. All \UTshort{} colleges were represented by auditors that enrolled in the course. Surveyed auditors reported less engagement in the course than students and some expressed having different expectations for the course.

\paragraph{Course participants appreciated the variety of speakers and connection to examples---the more the better!}
22\% of the 122 open-ended responses described the variety of guest speakers and broad overview. 
``I loved that we heard from different guest speakers every week; it was great to get to see so many experts in so many different fields. I also found the course content to be sufficiently reflective of the instructors' thesis that the AI problem is an interdisciplinary problem, not just a computer science problem as is commonly thought of.'' 
Some also highlighted value of connections to examples: ``Anytime the material was tied into real life examples or scenarios I found it very helpful and applicable. I also liked hearing from speakers with knowledge of different areas of AI.'' The lectures stood out as a useful learning tool with 37\% of open-ended responses mentioning them: ``I liked the lectures and content about how AI works at a fundamental level and what we need to be aware of as we start integrating its use in our real lives.''

\paragraph{The readings were the most challenging part of the course for our non-technical audience.}
In response to the open-ended survey item: ``What aspect(s) of the course needs improvement?'', most of the content-related comments (19\% of 115 open-ended responses) focused on the level and length of the reading: ``In my opinion, some of the reading was mathematically challenging, so I would recommend adding prerequisite courses for students.'' We provide our reading list online,\footnote{\url{https://www.cs.utexas.edu/~pstone/Courses/109fall23/assignments.html}} but based on the feedback recommend revising it to include more introductory materials.

Creating an engaged student community in a one-credit course was challenging.
Other course structure critiques (13\%) focused on interactivity and specific assignments. 
%: ``I would like to see an active student community, either making the lectures way more interactive, or implementing new and fun ideas for students to participate in.'' 
We address our plans for increased interactivity below.

% Take from \url{https://docs.google.com/document/d/1t1JeMZPTaidTGruuDScobHfN70be1tgleZj21lNdtQs/edit}
% or from \url{https://docs.google.com/document/d/1yAm3LmsRPRGwhdPCU1JWW0YIxnYB5HlQ0-hArwhfdTw/edit}

\section{Future Plans}

%Fall 2024 course

Based on the feedback from this first version of the course, we further developed it to a 3-credit-hour course, first offered in fall of 2024\footnote{\url{ https://www.cs.utexas.edu/~pstone/Courses/309fall24/}}. The expanded course covers the same set of topics and remains geared toward a broad audience, with no technical background. Due to the challenges in satisfying the expectations of both students and a broad pool of auditors, we designed this course exclusively for students.

We examined the readings from the first iteration of the course, with non-computer science course staff reviewing
them and making recommendations for modification and/or replacement. We replaced several technical
readings from journals with approachable blogs and popular news articles. The readings were
read and discussed by groups of students using a collaborative annotation tool to foster
asynchronous discussion and a sense of community.

The three-credit course includes weekly asyncrhonous modules with readings, videos, and assignments before an additional synchronous session each week. During the synchronous class, the instructor provides additional context, and leads discussions about the material. Each student attends class in the studio at least once during the semester, serving as the spokesperson for a group of remote students with whom they engage in discussion through an online chat. A series of (non-programming) assignments familiarizes students with the capabilities and limitations of some AI-based tools and technologies. A significant writing assignment focused on the societal and ethical implications of AI guides students to consider its real-world applications. Accordingly, the course satisfies an ethics requirement for \UTshort{}.  
% At this writing, we are gathering feedback and intend to report on it in the future.

\section{Acknowledgments}
This study was evaluated and considered exempt from the IRB of the University of Texas at Austin and was conducted in accordance with relevant ethical guidelines. This work was supported in part by the National Science Foundation (CAREER-2046955, NRT-2125858, FAIN-2019844), and UT Austin's Good Systems grand challenge. Any opinions, findings, and conclusions expressed in this material are those of
the authors and do not necessarily reflect the views of the sponsors. Peter Stone serves as the Chief Scientist of Sony AI and receives financial compensation for that role.
The terms of this
arrangement have been reviewed and approved by the University of Texas at Austin in accordance with
its policy on objectivity in research.

\bibliography{references}

\begin{thebibliography}{15}
\providecommand{\natexlab}[1]{#1}

\bibitem[{Auerbach and Silverstein(2003)}]{auerbach2003qualitative}
Auerbach, C.; and Silverstein, L.~B. 2003.
\newblock \emph{Qualitative data: An introduction to coding and analysis},
  volume~21.

\bibitem[{{Carnegie Mellon University}(2024)}]{cmu-ai}
{Carnegie Mellon University}. 2024.
\newblock B.S. in Artificial Intelligence.
\newblock \url{https://www.cs.cmu.edu/bs-in-artificial-intelligence/}.
\newblock Accessed: 2024-09-16.

\bibitem[{Geldhof et~al.(2018)Geldhof, Warner, Finders, Thogmartin, Clark, and
  Longway}]{geldhof2018revisiting}
Geldhof, G.~J.; Warner, D.~A.; Finders, J.~K.; Thogmartin, A.~A.; Clark, A.;
  and Longway, K.~A. 2018.
\newblock Revisiting the utility of retrospective pre-post designs: the need
  for mixed-method pilot data.
\newblock \emph{Evaluation and program planning}, 83--89.

\bibitem[{Howard and Dailey(1979)}]{howard1979response}
Howard, G.~S.; and Dailey, P.~R. 1979.
\newblock Response-shift bias: A source of contamination of self-report
  measures.
\newblock \emph{Journal of Applied Psychology}, 64(2): 144.

\bibitem[{Kong, Cheung, and Zhang(2021)}]{kong2021evaluation}
Kong, S.-C.; Cheung, W. M.-Y.; and Zhang, G. 2021.
\newblock Evaluation of an artificial intelligence literacy course for
  university students with diverse study backgrounds.
\newblock \emph{Computers and Education: Artificial Intelligence}.

\bibitem[{Littman et~al.(2022)Littman, Ajunwa, Berger, Boutilier, Currie,
  Doshi-Velez, Hadfield, Horowitz, Isbell, Kitano, Levy, Lyons, Mitchell, Shah,
  Sloman, Vallor, and Walsh}]{ai100}
Littman, M.~L.; Ajunwa, I.; Berger, G.; Boutilier, C.; Currie, M.; Doshi-Velez,
  F.; Hadfield, G.; Horowitz, M.~C.; Isbell, C.; Kitano, H.; Levy, K.; Lyons,
  T.; Mitchell, M.; Shah, J.; Sloman, S.; Vallor, S.; and Walsh, T. 2022.
\newblock Gathering Strength, Gathering Storms: The One Hundred Year Study on
  Artificial Intelligence (AI100) 2021 Study Panel Report.
\newblock arXiv:2210.15767.

\bibitem[{Ng et~al.(2023)Ng, Lee, Tan, Hu, Downie, and Chu}]{ng2023review}
Ng, D. T.~K.; Lee, M.; Tan, R. J.~Y.; Hu, X.; Downie, J.~S.; and Chu, S. K.~W.
  2023.
\newblock A review of AI teaching and learning from 2000 to 2020.
\newblock \emph{Education and Information Technologies}, 28(7): 8445--8501.

\bibitem[{Ng et~al.(2021{\natexlab{a}})Ng, Leung, Chu, and Qiao}]{ng2021ai}
Ng, D. T.~K.; Leung, J. K.~L.; Chu, K. W.~S.; and Qiao, M.~S.
  2021{\natexlab{a}}.
\newblock AI literacy: Definition, teaching, evaluation and ethical issues.
\newblock \emph{Proceedings of the Association for Information Science and
  Technology}, 58(1): 504--509.

\bibitem[{Ng et~al.(2021{\natexlab{b}})Ng, Leung, Chu, and
  Qiao}]{ng2021conceptualizing}
Ng, D. T.~K.; Leung, J. K.~L.; Chu, S. K.~W.; and Qiao, M.~S.
  2021{\natexlab{b}}.
\newblock Conceptualizing AI literacy: An exploratory review.
\newblock \emph{Computers and Education: Artificial Intelligence}.

\bibitem[{Salda{\~n}a(2021)}]{saldana2021coding}
Salda{\~n}a, J. 2021.
\newblock The coding manual for qualitative researchers.

\bibitem[{Saltz et~al.(2019)Saltz, Skirpan, Fiesler, Gorelick, Yeh, Heckman,
  Dewar, and Beard}]{ai_ethics_course_survey}
Saltz, J.; Skirpan, M.; Fiesler, C.; Gorelick, M.; Yeh, T.; Heckman, R.; Dewar,
  N.; and Beard, N. 2019.
\newblock Integrating Ethics within Machine Learning Courses.
\newblock \emph{ACM Trans. Comput. Educ.}, 19(4).

\bibitem[{Scassellati(2023)}]{yale-ai}
Scassellati, B. 2023.
\newblock AI for Future Presidents (CS170).
\newblock \url{https://zoo.cs.yale.edu/dsac/blog/2023/12/19/cpsc-170/}.
\newblock Accessed: 2024-09-16.

\bibitem[{{University of Texas at Austin}(2024)}]{utexas_ms_ai}
{University of Texas at Austin}. 2024.
\newblock M.S. in Artificial Intelligence.
\newblock \url{https://cdso.utexas.edu/msai}.
\newblock Accessed: 2024-09-16.

\bibitem[{Vekhter and Biswas(2023)}]{vekhter2023responsible}
Vekhter, J.; and Biswas, J. 2023.
\newblock Responsible robotics: a socio-ethical addition to robotics courses.
\newblock In \emph{Proceedings of the AAAI Conference on Artificial
  Intelligence}, volume~37, 15877--15885.

\bibitem[{Williams(2023)}]{williams2023review}
Williams, R. 2023.
\newblock {A Review of Assessments in K-12 AI Literacy Curricula}.
\newblock
  \url{https://randi-c-dubs.github.io/K12-AI-ed/Constructionist_AI_Assessments.pdf}.
\newblock Accessed: 2024-09-16.

\end{thebibliography}

\end{document}